\documentclass{article}
   
\title{Generative Latent Neural PDE Solver using Flow Matching}

\usepackage{lipsum}
\usepackage[utf8]{inputenc}
\usepackage[T1]{fontenc}    
\usepackage[bookmarks=true, colorlinks=true,linkcolor=blue!70,citecolor=green!80!black]{hyperref}
\usepackage[backend=biber,,natbib=true,style=nature]{biblatex}

\usepackage{wrapfig}
\usepackage{booktabs}  
\usepackage{longtable}
\usepackage{ducksay}
\usepackage{multicol}
\usepackage{etaremune}
\usepackage{afterpage}
\usepackage{capt-of}
\usepackage[table, x11names]{xcolor}
\usepackage{decorule}
\usepackage{scalerel,xparse}
\usepackage{enumitem} 
\usepackage[top=25mm, bottom=25mm, left=25mm, right=25mm,
            foot=5mm, marginparsep=1mm]{geometry}
\usepackage{tabularray}
\usepackage{multirow}
\usepackage{anyfontsize}
\usepackage{subcaption}
\usepackage[symbol*]{footmisc}

\usepackage{chngcntr}

\usepackage{graphicx}
\usepackage{tikz}
\usepackage{tikz-cd}
\usepackage{hf-tikz}
\usepackage{pgfplots} 
\pgfplotsset{compat=newest} 
\pgfplotsset{
        table/search path={figures/drawings},
    }
\usetikzlibrary{fadings}
\usetikzlibrary{shapes, arrows, fit, backgrounds, arrows.meta}

\usetikzlibrary{matrix}
\usetikzlibrary{shadows.blur}
\usetikzlibrary{patterns, tikzmark}
\usetikzlibrary{decorations.pathreplacing, calc, decorations.markings,}
\usetikzlibrary{positioning}
\usetikzlibrary{shapes.geometric}
\pgfdeclareplotmark{mystar}{
    \node[star,star point ratio=2.25,minimum size=5pt,
          inner sep=0pt,draw=black,solid,fill=green] {};
}

\usepgfplotslibrary{groupplots}
\usepgfplotslibrary{patchplots}

\definecolor{bg}{gray}{0.97}
\definecolor{olive}{rgb}{0.6, 0.6, 0.2}
\definecolor{sand}{rgb}{0.8666666666666667, 0.8, 0.4666666666666667}
\definecolor{wine}{rgb}{0.5333333333333333, 0.13333333333333333, 0.3333333333333333}
\definecolor{deblue}{RGB}{11,132,147}
\definecolor{ocra}{RGB}{204, 119, 34}

\usepackage{lettrine}
\usepackage{Typocaps}

\LettrineTextFont{\itshape}
\usepackage{amsthm}

\usepackage{amssymb}
\usepackage{pifont}
\usepackage{minitoc}

\newcommand{\chapref}[1]{\hyperref[#1]{Chapter \ref{#1}}}
\newcommand{\secref}[1]{\hyperref[#1]{Section \ref{#1}}}

\usepackage{marginnote}

\usepackage[many]{tcolorbox}
\tcbuselibrary{breakable,xparse,skins}

\usepackage{newunicodechar}
\newunicodechar{λ}{{$\mathtt\lambda$}}
\newunicodechar{μ}{{$\mathtt\mu$}}

\usepackage{xparse}
\DeclareTColorBox{emphbox}{O{black}O{0cm}}{
    enhanced jigsaw,
    breakable,
    outer arc=0pt,
    arc=0pt,
    colback=white,
    rightrule=0pt,
    toprule=0pt,
    top=0pt,
    right=0pt,
    bottom=0pt,
    bottomrule=0pt,
    colframe=#1,
    enlarge left by=#2,
    width=\linewidth-#2,
}

\DeclareTColorBox{emphbox}{O{black}O{0cm}}{
    empty,
    breakable=true,
    outer arc=0pt,
    arc=0pt,
    rightrule=0pt,
    leftrule=2pt,
    borderline west={2pt}{0pt}{#1},
    toprule=0pt,
    top=0pt,
    right=-3pt,
    bottom=0pt,
    bottomrule=0pt,
    colframe=#1,
    enlarge left by=#2,
    width=\linewidth-#2,
}

\usepackage{amsmath, amssymb, amsfonts, mathtools}
\usepackage{physics}
\usepackage{cancel}
\usepackage{thmtools, thm-restate}
\usepackage{accents}
\usepackage{bm}

\makeatletter
\newcommand{\ostar}{\mathbin{\mathpalette\make@circled *}}
\newcommand{\make@circled}[2]{%
  \ooalign{$\m@th#1\smallbigcirc{#1}$\cr\hidewidth$\m@th#1#2$\hidewidth\cr}%
}
\newcommand{\smallbigcirc}[1]{%
  \vcenter{\hbox{\scalebox{0.77778}{$\m@th#1\bigcirc$}}}%
}
\makeatother

\usepackage{pict2e, picture}
\makeatletter
\DeclareRobustCommand{\Arrow}[1][]{%
\check@mathfonts
\if\relax\detokenize{#1}\relax
\settowidth{\dimen@}{$\m@th\rightarrow$}%
\else
\setlength{\dimen@}{#1}%
\fi
\sbox\z@{\usefont{U}{lasy}{m}{n}\symbol{41}}%
\begin{picture}(\dimen@,\ht\z@)
\roundcap
\put(\dimexpr\dimen@-.7\wd\z@,0){\usebox\z@}
\put(0,\fontdimen22\textfont2){\line(1,0){\dimen@}}
\end{picture}%
}
\makeatother

\DeclareMathAlphabet{\nummathbb}{U}{BOONDOX-ds}{m}{n}

\makeatletter
\DeclareRobustCommand\widecheck[1]{{\mathpalette\@widecheck{#1}}}
\def\@widecheck#1#2{%
    \setbox\z@\hbox{\m@th$#1#2$}%
    \setbox\tw@\hbox{\m@th$#1%
       \widehat{%
          \vrule\@width\z@\@height\ht\z@
          \vrule\@height\z@\@width\wd\z@}$}%
    \dp\tw@-\ht\z@
    \@tempdima\ht\z@ \advance\@tempdima2\ht\tw@ \divide\@tempdima\thr@@
    \setbox\tw@\hbox{%
       \raise\@tempdima\hbox{\scalebox{1}[-1]{\lower\@tempdima\box
\tw@}}}%
    {\ooalign{\box\tw@ \cr \box\z@}}}
\makeatother

\bibliography{_bibliography/diffusion_models}
\bibliography{_bibliography/neural_operator}
\bibliography{_bibliography/misc}
\bibliography{_bibliography/gnn}

\begin{document}
 \author{%
  Zijie Li$^{\dagger}$, Anthony Zhou$^\dagger$, Amir Barati Farimani$^{\dagger}$\footnote{Correspondence: \texttt{barati@cmu.edu}}\\ 
  \textit{$^\dagger$ Department of Mechanical Engineering, Carnegie Mellon University}}
  \date{}
\maketitle
 \begin{abstract}
Autoregressive next-step prediction models have become the \textit{de-facto} standard for building data-driven neural solvers to forecast time-dependent partial differential equations (PDEs). Denoise training that is closely related to diffusion probabilistic model has been shown to enhance the temporal stability of neural solvers, while its stochastic inference mechanism enables ensemble predictions and uncertainty quantification. In principle, such training involves sampling a series of discretized diffusion timesteps during both training and inference, inevitably increasing computational overhead. In addition, most diffusion models apply isotropic Gaussian noise on structured, uniform grids, limiting their adaptability to irregular domains. We propose a latent diffusion model for PDE simulation that embeds the PDE state in a lower-dimensional latent space, which significantly reduces computational costs. Our framework uses an autoencoder to map different types of meshes onto a unified structured latent grid, capturing complex geometries. By analyzing common diffusion paths, we propose to use a coarsely sampled noise schedule from flow matching for both training and testing. Numerical experiments show that the proposed model outperforms several deterministic baselines in both accuracy and long-term stability, highlighting the potential of diffusion-based approaches for robust data-driven PDE learning.
\end{abstract}

\setlength\abovedisplayshortskip{2pt}
\setlength\belowdisplayshortskip{2pt}
\setlength\abovedisplayskip{2pt}
\setlength\belowdisplayskip{2pt}

\section{Introduction}

Data-driven neural solvers have emerged as promising surrogates for simulating partial differential equations (PDEs), offering remarkable speed-ups over traditional numerical methods. Despite their attractive properties, recent studies \citep{brandstetter2023mppde, li2022chaotic, lippe2023refiner, list2024temporalunroll} reveal that purely neural solutions often struggle with temporal stability, particularly in chaotic, time-dependent systems. Compounding errors frequently arise from the distribution shift between training and inference.  The distribution shift primarily stems from the spectral bias of neural network prediction \citep{rahaman2019spectralbiasneuralnetworks}, which causes inaccuracies in the modeling of high-frequency components. Although such errors might not be pronounced in the single-step prediction, they gradually propagate across the entire frequency spectrum over the course of the simulation.

A straightforward way to mitigate this problem is training perturbation, in which the model input is deliberately perturbed during training to emulate the distribution shift observed in inference. A common choice for perturbation is isotropic Gaussian noise \citep{sanchez2020gns, stachenfeld2022learned}, where the specific noise level typically depends on the problem at hand and requires tuning. Backpropagation Through Time (BPTT) \citep{bptt1990} is a promising alternative that has been shown to effectively improve the stability of neural solvers across a variety of physical systems \citep{lam2023graphcast, brandstetter2023mppde, list2024temporalunroll}. By unrolling the neural surrogate over multiple timesteps during training, it offers two practical advantages: (1) the model can see its own prediction which reduces the gap between training data and testing data, and (2) it is optimized on a multi-step loss that promotes long-term stability. Despite its effectiveness, BPTT substantially raises training costs by requiring model simulations and storing intermediate activations over the unrolled trajectory. In addition, extending backpropagation over many timesteps can lead to unstable gradients, complicating the optimization. Truncating the backpropagation chain, for example, using the pushforward technique proposed in \citet{brandstetter2023mppde}, can reduce computational demands. Nevertheless, because simulation is still required during training, its overall cost remains higher than that of the standard next-step regression.

Recently, the success of diffusion probabilistic models \citep{sohl2015unsupervised, ho2020denoising, song2021scorebased} in image generation has sparked a rapid expansion of their application to PDE-related problems \citep{lippe2023refiner, zhou2025text2pde, shu2023physics, kohl2024acdm, huang2024diffusionpde, valencia2025learning, price2024gencast, du2024confild}. The denoising training target and multi-step sampling are found to alleviate the spectral bias of neural network prediction, which is particularly important for achieving stable long-term prediction in PDE simulation. In particular, the training of diffusion models are simulation-free, which are much more efficient than BPTT. The probabilistic nature of diffusion models also opens up a new venue for approaching the simulation of highly chaotic systems like 3D turbulence \citep{lienen2024zeroturb}. Achieving accurate long-term prediction on turbulent systems often requires resolving small-scale physics that demands very fine spatio-temporal discretization, and even a very minor error made in early prediction can result in significantly different future states. Instead of reproducing the exact individual trajectory, diffusion models can generate diverse possible states that remain statistically consistent.

Although diffusion models show promise for predicting and simulating time-dependent PDEs, they also pose notable challenges. In particular, their multi-step characteristic significantly raises computational costs during inference. Moreover, training these models can be computationally intensive due to slow convergence \citep{hang2024minsnr}: they rely on Monte Carlo sampling of diffusion time steps to approximate the integral in the evidence lower bound \citep{ho2020denoising} or the denoise score matching objective \citep{vincent2011denoise, song2021scorebased}. In this work, we propose to address these challenges from two complementary perspectives. First, we analyze different diffusion paths commonly used in generative image modeling, including the Denoising Diffusion Probabilistic Model (DDPM) \citep{ho2020denoising} and more recent flow matching approaches \citep{lipman2023flow, liu2023flow}, and we find that a coarsely truncated noise schedule empirically works well. This reduces the number of diffusion training steps needed for both training and sampling. Second, we propose to do diffusion in a unified mesh-reduced latent space, leveraging an autoencoder that is designed to map functions sampled on disparate, irregular meshes into a structured compressed latent space. This dimensionality reduction enables a streamlined pipeline for training and deploying diffusion-based neural PDE solvers more efficiently. In addition, by trading off a certain degree of local (per-frame) accuracy, we achieve more stable predictions in the mesh-reduced space, a concept somewhat analogous to pseudo-spectral methods, where high-frequency modes are filtered to balance global stability and local accuracy.

\section{Related Works}

\paragraph{Neural PDE solver} 

Neural networks can learn and approximate solution mappings in PDE problems directly from data \citep{azizzadenesheli2024neural, hao2022physics, karniadakis2021physics}, utilizing mesh-specific architectures such as convolutional layers for uniform grids \citep{stachenfeld2022learned, guo2016cnnfluid, tompson2017cnnfluid} or graph-based layers for unstructured meshes \citep{sanchez2020gns, pfaff2021mgn, li2022gamd, li2022fgn, lino2022multiscale, brandstetter2023mppde}. Neural operators \citep{lu2021deeponet, kovachki2023neural} aim to learn a continuous mapping between input and target functions for a family of PDEs, potentially allowing them to use a single set of parameters across varying discretizations. The practical realization of it includes the learned transformation in different spectral spaces \citep{li2021fno, gupta2021multiwavelet, fanaskov2024spectralno}, non-linear kernel parameterized with message passing graph neural network\citep{li2023gino, li2020multipolegno} or kernel parameterized with dot-product attention \citep{gnot2023icml, wu2024transolver, factformer2023nips, oformer2023tmlr, cao2021galerkin}. In addition, neural network can also be used to parameterize the solution function of a single instance of PDE and optimized in a data-free manner, which is known as Physics-Informed Neural Networks (PINNs) \citep{raissi2019physics}. The physics-informed loss can also be applied data-driven neural operator to improve accuracy and generalization \citep{li2023pino, wang2021pi-deeponet}.%
\paragraph{Neural solver on reduced-order representations} Reduced Order Modeling (ROM) \citep{benner2015romsurvey} is widely employed to build low-dimensional, computationally efficient representations of high-dimensional dynamical systems. The reduced latent vector can be obtained from data through linear projection, e.g. proper orthogonal decomposition \citep{berkooz1993proper}, non-linear projection methods like neural-network-based autoencoder \citep{Lui2019nnrom, murata2020nonlinearrom, maulik2021reduced, fresca2021poddlrom} or learned basis function \citep{pan2023nif, chen2022crom, serrano2023coral}. The temporal dynamics can be linearized and modeled with Koopman operator \citep{brunton2021modernkoopman, Deep-Koopman-Nature-2018}, or directly modeled with neural networks \citep{yin2023continuouspde, rojas2021reduced, hemmasian2023reduced, vlachas2022led, wiewel2019latentspacephysics, kontolati2023latentdeeponet}. Reducing the dimensionality of the discretized vector significantly lowers the training cost of a neural network operating in the reduced space, which enables longer BPTT horizon for more stable dynamics forecasting \citep{han2022predicting, li2025lns}.

\paragraph{Temporal stability of neural solver} For numerical solvers that employ an explicit scheme, Courant-Friedrichs-Lewy (CFL) condition poses a constraint on temporal discretization for controlling the propagation of local approximation error. For neural solvers, they are not restricted from such constraints but the propagation of the local error are often unbounded. Common remedies include training perturbation (e.g. noise injection) \citep{sanchez2020gns}, BPTT with unrolled training \citep{list2024temporalunroll, lam2023graphcast, han2022predicting} and truncated BPTT \citep{brandstetter2023mppde}. Empirical evidence indicates that aliasing-related spectrum artifacts substantially contribute to rollout instability in certain neural network architectures \citep{worrall2025spectral, mccabe2023towards, raonic2023convolutional}. Techniques like low-pass filtering have been shown to suppress these artifacts and improve stability. Neural solvers can also be improved through physics-guided post-processing. For instance, \citet{li2022chaotic} shows that enforcing dissipativity on the network’s predictions enhances long-term statistical accuracy. \citet{cao2025spectralrefiner} proposes a spectral fine-tuning method for fine-tuning neural solver with functional-type norm to minimize the residual that reaches accuracy comparable to numerical solvers on temporal forecasting of 2D fluid flow while being highly efficient.

\paragraph{Diffusion probabilistic models in PDEs} The success of diffusion models in image generation \citep{ho2020denoising, rombach2022ldm} has spurred growing interest in applying diffusion model for PDE problems. \citet{shu2023physics} proposes to incorporate the constraint of the solution manifold \citep{chung2024dps}, i.e. the residual information of the PDEs, into the posterior sampling process for solving inverse problem in fluid field reconstruction. \citet{lippe2023refiner} find that denoising training target can effectively alleviate spectral bias in neural network prediction and propose a modified noise schedule for denoising training/sampling. Similar observation is made in autoregressive conditional diffusion model \citep{kohl2024acdm}. The conditional diffusion framework has also been applied in global weather forecasting \citep{price2024gencast}, not only it improves the stability and accuracy of the forecast, but also provides uncertainty quantification of extreme event. Diffusion model can also be used for data assimilation \citep{huang2024diffusionpde, shysheya2024conditional}, uncertainty quantification \citep{shu2024zeroshotuq, liu2024airfoiluq} and learning statistically coherent distribution \citep{valencia2025learning, lienen2024zeroturb, li2024syntheticturb} in PDE problems.

\section{Methodology}

\subsection{Diffusion probabilistic model}
Diffusion probabilistic models (DPMs) \citep{sohl2015unsupervised, song2021scorebased, ho2020denoising} are a class of generative models that reverse noise-corruption processes. Given data point $\mathbf{x}_0 \in \mathbb{R}^{d}$ and its associated distribution $\mathbf{x}_0  \sim p(\mathbf{x}_0)$, (Gaussian-) diffusion model defines a forward process that gradually adds Gaussian noise to the state variable $\mathbf{x}_t$ which results in following time-dependent marginal: $q(\mathbf{x}_t|\mathbf{x}_0) = q(\mathbf{x}_t|\alpha_t \mathbf{x}_0, \sigma_t^2 \mathbf{I})$, where $\alpha_t, \sigma_t \in \mathbb{R}_+$ are hyperparameters of the noising process, and are usually referred to as noising schedule. The noise schedule is chosen such that the endpoint $q(\mathbf{x}_T | \mathbf{x}_0) \approx \mathcal{N}(0, \bar \sigma^2 \mathbf{I})$. 

It has been shown that the following stochastic differential equation (SDE) result in the above transition distribution $q(\mathbf{x}_t|\mathbf{x}_0)$  \citep{song2021scorebased}:
\begin{equation}
    d \mathbf{x}_t = f(t) dt + g(t) d\mathbf{w}_t,
    \label{eq:diffusion forward sde}
\end{equation}
where $\mathbf{w}_t \in \mathbb{R}^d$ is the standard Wiener process, and 
\begin{equation}
    f(t) = \frac{d\log \alpha_t}{dt}, g^2(t) = \frac{d\sigma_t^2 }{dt} - 2 \frac{d\log \alpha_t}{dt}\sigma_t^2.
    \label{eq:diffusion coefficient}
\end{equation} \citet{song2021scorebased} also show that the SDE in \eqref{eq:diffusion forward sde} can be reversed by solving the following SDE backward in time:
\begin{equation}
    d \mathbf{x}_t = [f(t) \mathbf{x}_t - g^2(t)\nabla_x \log q_t(\mathbf{x}_t)]dt + g(t)d\bar{\mathbf{w}}_t,
    \label{eq:diffusion backward sde}
\end{equation}
where $\bar{\mathbf{w}}_t$ is the standard Wiener process flowing reverse in time. Furthermore, there also exists an ordinary differential equation (ODE) which shares the same marginal probabilities along the trajectory $\{q_t(\mathbf{x}_t)\}$:
\begin{equation}
    d \mathbf{x}_t = [f(t)-\frac{1}{2}g^2(t)\nabla_x \log q_t(\mathbf{x}_t)]dt.
    \label{eq:probability flow ode}
\end{equation}
The score function $\nabla_x \log q_t(\mathbf{x}_t)$ can be estimated with a neural network model $s_\theta$ that is trained via denoising score matching \citep{vincent2011denoise, song2021scorebased, song2019generative}. To generate samples, one can first sample $\mathbf{x}_T \sim \mathcal{N}(0, \bar \sigma^2 \mathbf{I})$ and then numerically solve \eqref{eq:diffusion backward sde} or \eqref{eq:probability flow ode} backward in time. The ODE formulation is generally more tolerant to larger step size and thus it is widely adopted for faster sampling \citep{song2021ddim, lu2022dpm, zhang2023deis}. 

The conditional information $\mathbf{y}$ can be plugged into the model directly to estimate the conditional score function $\nabla_x \log q_t(\mathbf{x}_t | \mathbf{y})$. 

For forward problems in time-dependent PDEs, we are interested in estimating $\mathbf{u}^{m+1}$ given: $\mathbf{u}^m, \mathbf{u}^{m-1}, \hdots$, $\mathbf{u}^{m-h+1}$, $\xi$, where $\mathbf{u}^{m}$ denotes the system state $\mathbf{u}$ (e.g. velocity, pressure) at physical timestep $t_m$, and $\xi$ denotes system parameters such as viscosities. The most straightforward way to achieve this would be training a regression model to predict $\mathbf{u}^{m+1}$. However, it is empirically observed that the long-term rollout stability for this kind of model is usually brittle despite good next-step prediction accuracy \citep{brandstetter2023mppde}. The major reason for this is the input distribution shift due to the accumulated error of the model. The discrepancy between model's next-step prediction and ground truth in the high-frequency regime will gradually transport to the whole spectrum through the non-linear terms in the equations (e.g. $u\nabla u$), even though the error in this regime might not be pronounced in loss function like MSE. PDE-Refiner \citep{lippe2023refiner} shows that recasting next-step predictor as an interpolator between noise and next-step prediction can mitigate the spectral bias of neural network's prediction \citep{rahaman2019spectralbiasneuralnetworks} and therefore improve the long-term stability. ACDM (autoregressive conditional diffusion model) \citep{kohl2024acdm} reformulates deterministic next-step prediction as probabilistic generation of the future state, and observes a similar trend as PDE-refiner. Despite these models are built from different perspectives but they are actually equivalent under the following viewpoint, where they are just different parameterizations of an interpolant between noise $\mathbf{x}_1=\bar{\mathbf{\epsilon}} \sim \mathcal{N}(0, \mathbf{I}) $ and next-step state $\mathbf{x}_0 = \mathbf{u}^{m+1}$. A commonly used sampler/numerical solver for the diffusion model is DDIM \citep{song2021ddim}, with the following update rule:
\begin{equation}
    \mathbf{x}_s = \frac{\alpha_s}{\alpha_k} \mathbf{x}_k  - \alpha_s [\frac{\sigma_k}{\alpha_k} - \frac{\sigma_s}{\alpha_s}]\hat{\mathbf{\epsilon}}_k,
\end{equation}
where $0 < s < k $, $\hat{\mathbf{\epsilon}}$ denotes the noise prediction of the neural network. Under change of variable, $\lambda_i = \sigma_i / \alpha_i, \tilde{\mathbf{x}}_i = \mathbf{x}_i / \alpha_i$:
\begin{equation}
    \tilde{\mathbf{x}}_s = \tilde{\mathbf{x}}_k - (\lambda_k - \lambda_s) \hat{\mathbf{\epsilon}}_k.
    \label{eq:ddim one step}
\end{equation}

For the SDE formulation (as the ancestral sampler in DDPM \citep{ho2020denoising} and PDE-Refiner \citep{lippe2023refiner}) amounts to the Euler scheme in equation \eqref{eq:ddim one step} with additional noise injection at every update: 
\begin{equation}
\tilde{\mathbf{x}}_s = \tilde{\mathbf{x}}_k - 2(\lambda_k - \lambda_s) \hat{\mathbf{\epsilon}}_k + \sqrt{\lambda_k^2 - \lambda_s^2} \mathbf{n}_k,
\end{equation}
where $\mathbf{n}_k \sim \mathcal{N}(0, \mathbf{I})$. Given that $\lambda(i)=\sigma(i)/\alpha(i)$ is monotonic, \eqref{eq:ddim one step} is the first-order Euler discretization of the following ODE:
\begin{equation}
    d\tilde{\mathbf{x}}_{\lambda} = - \mathbf{\epsilon}(\tilde{\mathbf{x}}_{\lambda}, \lambda)d\lambda.
    \label{eq:dpm change of variable}
\end{equation}

Under the context of next-step prediction in temporal PDEs, if we use diffusion integration scheme to solve for the next step, considering the DDIM update in \eqref{eq:ddim one step}, the final prediction is a linear combination of a series of prediction:

\begin{equation}
\begin{aligned}
    \tilde{\mathbf{x}}_{\lambda_0} &= \tilde{\mathbf{x}}_{\lambda_1} -\hat{\mathbf{\epsilon}}_{\lambda_1}  (\lambda_1 - \lambda_0) \\
    &= \left(\tilde{\mathbf{x}}_{\lambda_2} - \hat{\mathbf{\epsilon}}_{\lambda_2}   (\lambda_2 - \lambda_1)\right)-\hat{\mathbf{\epsilon}}_{\lambda_1}  (\lambda_1 - \lambda_0) \\
    &= \tilde{\mathbf{x}}_{\lambda_K} - \sum_{i=1}^{K} \hat{\epsilon}_{\lambda_i}(\lambda_i - \lambda_{i-1}), 
    \label{eq:diffusion multi-step}
\end{aligned}
\end{equation}
where the diffusion temporal discretization $\{\lambda_i\}_{i=0}^K$ determines the coefficients for each neural network prediction. Interestingly, this is in the same functional form as many linear multi-step methods, such as Adams–Bashforth method. The "diffusion multi-step" method here offers several intriguing properties when building a neural PDE solver. First, it can be viewed as predictor-corrector scheme that allows the prediction error made at earlier diffusion steps being adjusted at later diffusion steps. From \eqref{eq:diffusion multi-step}, the final weight of model can be dynamically adjusted with noise schedule $\alpha_i, \sigma_i$. Second is that training model different levels of noise facilitates the optimization of model's prediction at different regimes of the spectrum. Furthermore, the diffusion-based prediction effectively introduces stochasticity to the model. It is extremely difficult to obtain accurate long-term forecast for a lot of the chaotic systems that are sensitive to small perturbations. In such case, deterministic forecasting models will produce either blurry prediction or only be able to generate a single set of system state, whereas stochastic models can provide a better coverage of different modes in the distribution of possible system states.


The noise schedule is a critical factor that controls both the training and inference dynamics. ACDM \citep{kohl2024acdm} have explored using DDPM's scheduler \citep{ho2020denoising}, which works well on image generation tasks, to temporal PDE forecasting. It is observed that the backward diffusion sampling converges within fewer DDIM steps than natural image generation. This is not surprising as for a well-determined forward problem, the distribution $p(\mathbf{u}^{m+1}|\mathbf{u}^m)$ is a Dirac delta function centered around the unique solution of $\mathbf{u}^{m+1}$. In contrast, conditional image generation is a many-to-one mapping problem (i.e. multiple text description can correspond to the same image and vice versa), such that too few sampling steps will result in sampling the mean of multiple possible images that is usually blurry. Therefore, forward problems in PDEs might not require much of the inference steps, and even training steps.
To this end, PDE-Refiner \citep{lippe2023refiner} proposes a customized exponential scheduler and shows that it outperforms the DDPM scheduler. However, it requires heuristically picking a minimum noise level for the noise schedules and additionally, $\lambda_i=\alpha_i/\sigma_i$ are more shifted towards high signal-to-noise regime which might cause the input samples not perturbed enough.
In this work, we propose to develop our generative latent neural PDE sovler based on flow matching framework. Flow matching \citep{lipman2023flow, liu2023flow} with optimal transport path and Gaussian endpoint $q_1(\mathbf{x}_1) \sim \mathcal{N}(0, \mathbf{I}) $, can also be viewed as a special case of Gaussian diffusion with following noise schedule:
\begin{equation}
    \mathbf{x}_t = (1-t) \mathbf{x}_0 + t \mathbf{\epsilon},  \quad \mathbf{\epsilon} \sim \mathcal{N}(0, \mathbf{I}).
\end{equation}
Based on the above noise schedule, \citet{lipman2023flow, liu2023flow} propose the following flow ODE: $d \mathbf{x}_t  = -\mathbf{v} (\mathbf{x}_t, t)dt$, where a neural network $\mathbf{v}_\theta$ is trained to estimate the velocity field via optimizing the conditional flow matching loss \citep{lipman2023flow}. From \eqref{eq:dpm change of variable}, we can see that the flow matching with the optimal path shares a similar flow ODE as the standard diffusion probabilistic model up to a change of variable. Compared to the commonly used noise prediction parameterization, the velocity field prediction in flow matching target offers several practical benefits. As it avoids the irregularity at the boundary $t=1$ during sampling that arises due to the $\alpha_1$ in the denominator \citep{lin2024diffusionschedules}. Additionally, flow matching's noise schedule provides better coverage of low signal-to-noise ratio regime compared to the exponential schedule in PDE-Refiner, where we observe that it empirically performs better on mesh-reduced space.



\subsection{Diffusion in a unified mesh-reduced space}

\begin{figure}[t]
    \centering
    \includegraphics[width=\linewidth]{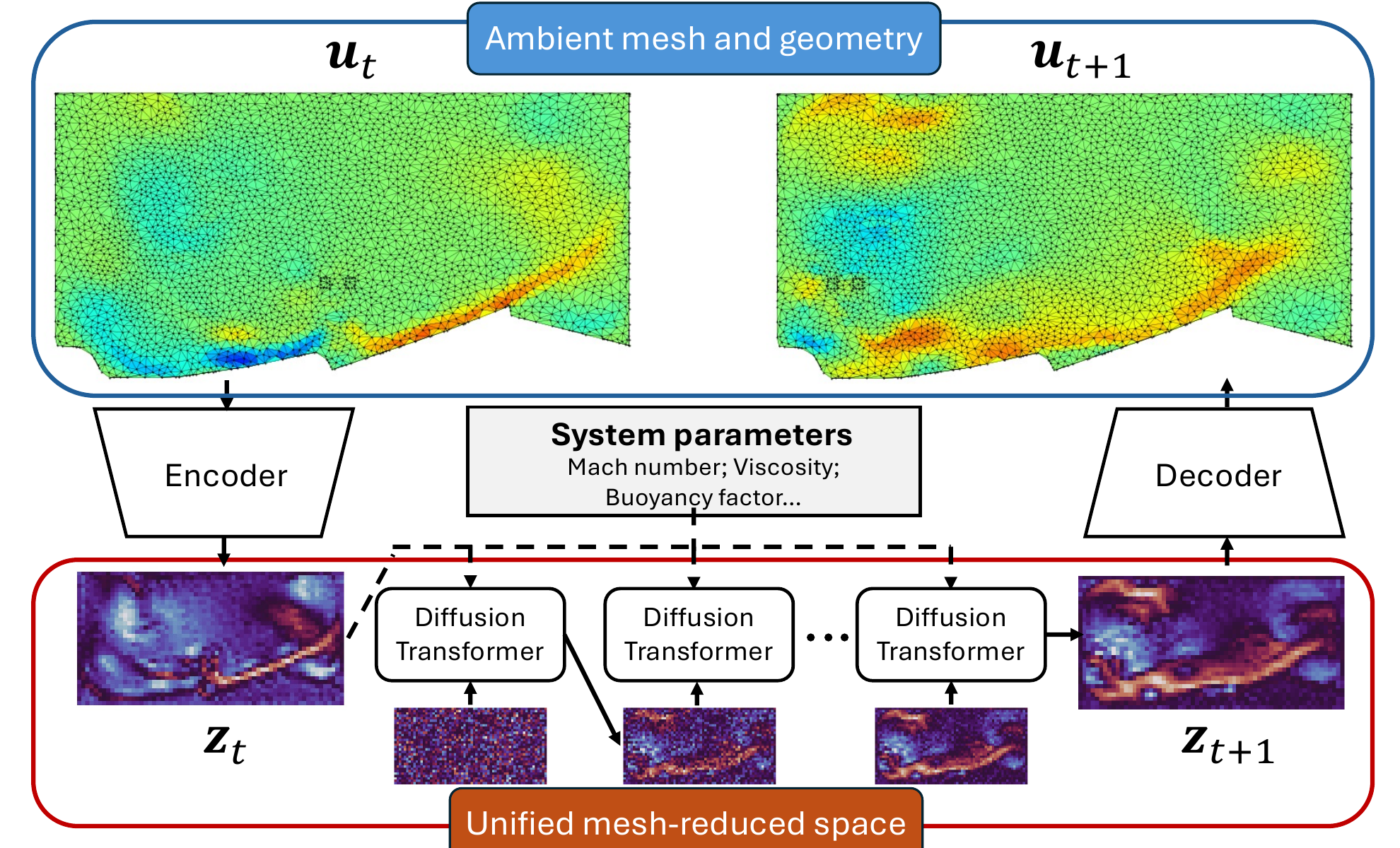}
    \caption{Main architecture of the proposed latent neural solver based on flow matching. The input data is projected onto a unified latent space via encoder. In the latent space, the diffusion model iteratively transform the noise into the latent state of future timestep. The latent embedding can be projected back to the original mesh space via the decoder.}
    \label{fig:main architecture}
\end{figure}

The multi-step inference scheme in diffusion models, while offering certain advantages in terms of accuracy and stability, makes them more computationally expensive than a standard next-step prediction model. Moreover, as many PDE data are often represented on irregular meshes with complex geometries, which can make the Gaussian diffusion formulation less effective. Inspired by the success of latent diffusion model in image modeling \citep{rombach2022ldm}, we propose to learn diffusion model in a latent space that features a resolution-reduced discretized field and uniform grid structure.

\paragraph{Autoencoder} The encoder is used to project the function sampled on input meshes to latent embedding that lies on a coarsened grid, and the decoder is used to reconstruct the target function values on a set of query points. To approximate the mapping between functions and transform the discretization structure of the field (more specifically, from an irregular mesh to a uniform mesh), we leverage the learnable kernel integral proposed in \citet{kovachki2023neural}, which serves as a well-suited foundational component. Specifically, we parameterize the kernel as a distance-based function and use locally truncated Monte-Carlo sampling to compute the integral numerically as in Geometry-Informed Neural Operator (GINO) \citep{li2023gino}:

\begin{equation}
    f(y_i) = \int_{B_r(y_i)} \kappa(y_i, y) u(y) dy \approx \sum_{j\in B_r(i)} \kappa(y_i, y_j) u(y_j) \mu_j,
    \label{eq:gino project}
\end{equation}
where $f$ is the output function and $u$ is the input function, $B_r$ denotes a local ball with radius $r$, $\mu_j$ is the Riemann sum weight for $j$-th point. \eqref{eq:gino project} only requires point-based sampling, thus it is mesh-agnostic and enables altering the discretization within the model. The truncated domain reduces the number of points that need to be taken into consideration when evaluating the integral, which greatly improves model's efficiency on high-dimensional problems. After deriving functions sampled on uniform latent grid, we use convolutional layers to further compress/reconstruct it. Despite convolutional layers are less flexible in terms of handling different grid resolutions, it is observed that they performs well on certain fixed-resolution problems \citep{zhou2025text2pde, gupta2022multispatiotemporal}, primarily due to that the numerical accuracy of evaluating an integral on coarse mesh can be limited.

The optimization target of the autoencoder is to achieve least lossy compression of the PDE data. In the meantime, since our ultimate goal is to develop an accurate and stable temporal PDE prediction model, we want to avoid a high variance and temporal oscillating latent space. To this end, the training target for the autoencoder comprises three components - an $\mathcal{L}^2$ reconstruction loss, a KL divergence regularization term \citep{kingma2022vae}
that pushes the latent distribution towards a Gaussian prior, and temporal jerk regularization \citep{xie2024jerkreg} that promotes the smoothness of latent dynamics.

\paragraph{Diffusion Transformer} 
We parameterize the velocity predictor (of the diffusion process) using an attention-based \citep{vaswani2023attention} neural network architecture called Diffusion Transformer. Diffusion Transformer (DiT) performs well in the diffusion modeling of text \citep{sahoo2024mdlm} and image \citep{peebles2023dit}. Given the context of previous frames' latent encoding $\{\mathbf{z}^{m}, \mathbf{z}^{m-1}, \hdots, \mathbf{z}^{m-h+1}\}$, system parameters $\xi$, a scalar indicator of the signal-to-noise ratio (here we use diffusion timestep $k$) and the state of prediction target at current diffusion timestep $\mathbf{x}_k$, the model $\mathbf{v}_\theta$ predicts the conditional velocity field $\mathbf{v}$ of the flow ODE: $\mathbf{v}(\mathbf{x}_k, k, \mathbf{c})=\mathbf{\epsilon} - \mathbf{x}_0$, where $\mathbf{x}_0=\mathbf{z}^{m+1}, \mathbf{x}_1=\epsilon \sim \mathcal{N}(0, \mathbf{I})$ and $\mathbf{c}$ contains the conditioning information of previous frames and system parameters. The loss function is defined as:
\begin{equation}
    L_{\text{FM}} = \mathbb{E}_{k \sim p(k), \mathbf{\epsilon} \sim \mathcal{N}(0, \mathbf{I})} || \mathbf{v}_\theta(\mathbf{x}_k, k, \mathbf{c}) - (\mathbf{\epsilon} - \mathbf{x}_0)  ||_2^2.
    \label{eq:diffusion training loss}
\end{equation}

To reduce the computational cost associated with data on multi-dimensional domain, we propose to replace part of the full-rank standard attention layers in DiT with multi-dimensional factorized attention proposed in \citet{factformer2023nips}, which we find not only improves efficiency but also the accuracy on the problems we studied.


\section{Experiments and Discussion}

\subsection{Problems}

We test out the proposed model on three different time-dependent PDE problems.  

\paragraph{2D buoyancy-driven flow} This problem depicts smoke volume rising in a closed domain. The governing equations are incompressible Navier-Stokes equation coupled with advection equation. The boundary condition for the smoke field is Dirichlet and the boundary condition for the flow field is Neumann. We use the pre-generated dataset from \citet{gupta2022multispatiotemporal} that is implemented in \textit{phiflow}\citep{holl2024phiflow}. The dataset contains 2496 training trajectories with varying initial conditions and buoyancy factors. The objective is to predict the scalar density field of smoke $d$ and velocity of flow $\mathbf{u}$.

\paragraph{Magnetohydrodynamics compressible turbulence} Magnetohydrodynamics (MHD) describes the behavior of electrically conducting fluids in the presence of magnetic fields and is fundamental to understanding astrophysical phenomena such as the solar wind, galaxy formation, and interstellar medium (ISM) turbulence \citep{burkhart2020mhd}. We use the publicly available dataset from the Well \citep{ohana2025well}. This dataset consists of isothermal MHD simulations in the compressible regime. The simulations, originally generated at a resolution of 
$256^3$, have been downsampled to $64^3$ using an ideal low-pass filter to remove aliasing artifacts. The dataset contains trajectories with different sonic Mach numbers and Alfvénic Mach numbers. The objective is to predict the density field $\rho$, magnetic field $\mathbf{B}$ and velocity field $\mathbf{u}$.

\paragraph{Airflow around Unmanned Aerial Vehicle} For this problem we use the EAGLE dataset \citep{janny2023eagle}. The EAGLE dataset simulates airflow dynamics around a 2D Unmanned Aerial Vehicle (UAV) hovering over different floor profiles. The UAV follows a fixed trajectory while its propellers create complex turbulence interacting with the scene. Floor profiles are generated using three interpolation methods: Step (sharp angles, abrupt flow changes), Triangular (small vortices), and Spline (smooth trails, complex vortices). The dataset includes 600 geometries and 1,200 simulation trajectories. Velocity and pressure fields are simulated on a dynamically evolving triangle mesh using ANSYS Fluent, solving Reynolds-Averaged Navier-Stokes equations with a Reynolds stress turbulence model. The target is to predict the pressure field $\mathbf{p}$ and velocity field $\mathbf{u}$.

\subsection{Discussion}

\begin{minipage}{0.46\textwidth}
\centering
\fontsize{10pt}{10pt}\selectfont
\setlength{\tabcolsep}{2pt}
\centering
\begin{tabular}{lcc} 
\toprule
Variables & $\mathbf{u}$ & d \\ 
\midrule
FNO \citep{li2021fno} & 0.5672 & 0.1308  \\
UNet \citep{ronneberger2015unet, gupta2022multispatiotemporal} & 0.4143 & 0.1059 \\ 
LDM (sequence) \citep{zhou2025text2pde} & 0.5384 & 0.2022 \\
\midrule
Latent-AR & 0.4229 & 0.1076 \\
Latent-FM (ens=1) & 0.3786 & 0.0995 \\
Latent-FM (ens=8) & \textbf{0.3759} & \textbf{0.0992} \\
\bottomrule
\end{tabular}
\captionof{table}[NRMSE of 2D buoyancy-driven flow]{NRMSE of different models on buoyancy-driven flow. \textit{ens} denotes the size of ensemble. \label{tab: nrmse 2d smoke}}
\end{minipage}
\begin{minipage}{0.52\textwidth}
\centering
\fontsize{9.5pt}{10pt}\selectfont
\setlength{\tabcolsep}{2pt}
\begin{tabular}{lcc} 
\toprule
Variable & $\mathbf{u}$ & $d$ \\ 
\midrule
Exponential ($\sigma_{\text{min}}$=1e-1) & 0.4837 & 0.1324 \\
Exponential ($\sigma_{\text{min}}$=1e-2) & 0.4289 & 0.1147 \\
Exponential ($\sigma_{\text{min}}$=1e-3) & 0.4023 & 0.1031 \\
Exponential ($\sigma_{\text{min}}$=1e-6) & 0.3995 & 0.1005 \\ 
\midrule
FM (5 steps) & 0.3859 & 0.1003 \\
FM (10 steps) & \textbf{0.3786} & \textbf{0.0995} \\
FM (1000 steps *) & 0.4514 & 0.1163 \\
\bottomrule
\end{tabular}
\captionof{table}[Diffusion ablation on 2D buoyancy-driven flow]{NRMSE of different diffusion paths on buoyancy-driven flow. *: 1000 steps are used for training and 50 steps are used for sampling. \label{tab: diffusion ablation}}
\end{minipage}
\vspace{+3mm}

On the buoyancy-driven flow problem, we consider benchmarking against FNO, modern UNet from \citet{gupta2022multispatiotemporal}, and sequence-wise latent diffusion model \citep{zhou2025text2pde}. We also implement a deterministic predictor which also operates in the latent space derived from autoencoder.
The result is listed in Table \ref{tab: nrmse 2d smoke}, which shows that the proposed flow matching predictor outperforms other deterministic baselines. The ensemble mean has slightly better accuracy over a single sampler. To study the influence of different noise schedules, we compare the performance of the model with different diffusion paths and training settings. For the exponential schedule proposed in PDE-Refiner, we alter the magnitude of $\sigma_{\text{min}}$ and compare their performance. All the exponential noise schedule based models use a sampling and training steps of 10. As for all flow matching (FM) based models, except for the 1000 steps variant, the training/sampling steps are indicated by the number in the bracket. As shown in Table \ref{tab: diffusion ablation}, flow matching outperforms the exponential schedule while eliminating the need to pre-define a minimum noise level. In addition, we observe that the discretization of diffusion timesteps during training can also be truncated, as a finer discretization of the diffusion time does not bring in performance improvement.
\begin{center}
  \begin{minipage}{0.49\textwidth}
\centering
        \includegraphics[width=\linewidth]{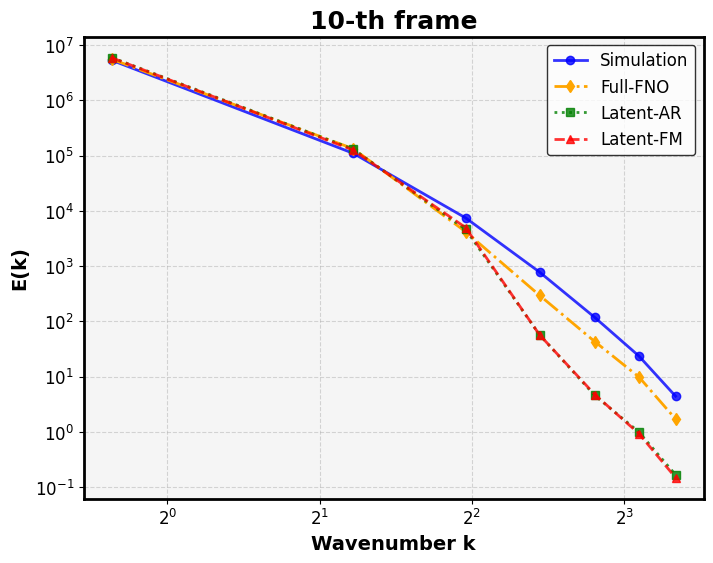}
        \captionof{figure}{Spectrum averaged across a time-window centered at 10-th frame.}
        \label{fig:spectrum 10th frame}
\end{minipage}
\begin{minipage}{0.49\textwidth}
\centering
        \includegraphics[width=\linewidth]{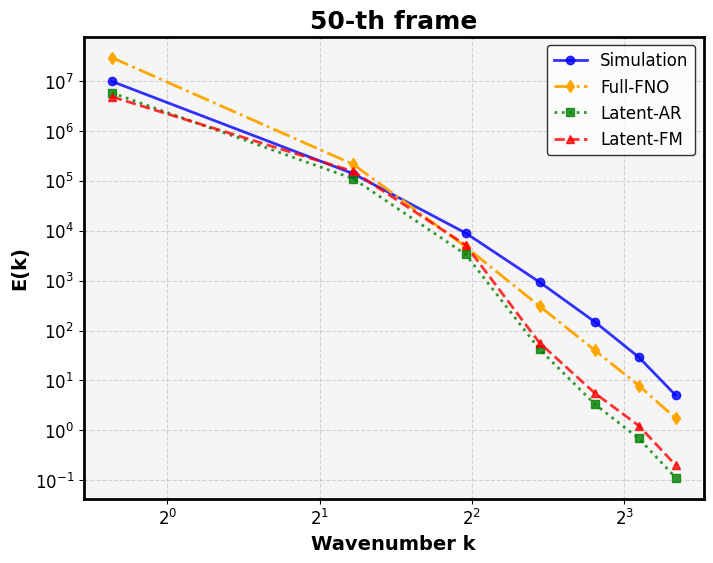}
        \captionof{figure}{Spectrum averaged across a time-window centered at 50-th frame.}
        \label{fig:spectrum 50th frame}

\end{minipage}
\begin{minipage}[T]{0.49\textwidth}
\centering
        \includegraphics[width=\linewidth]{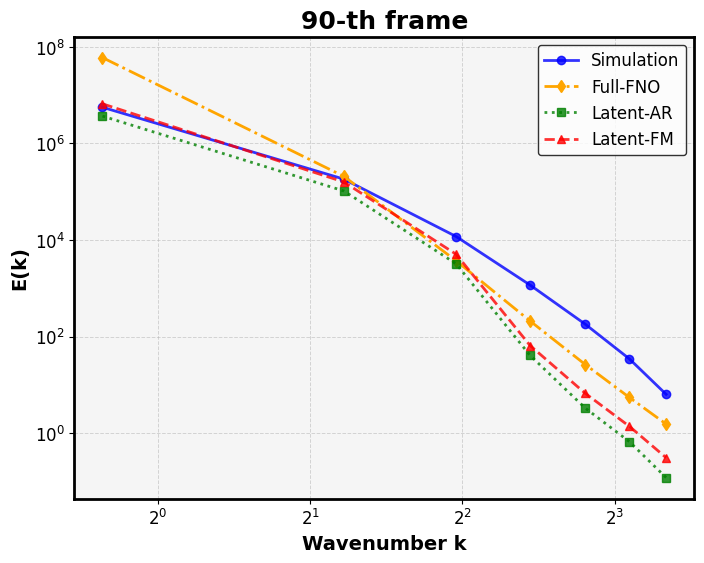}
        \captionof{figure}{Spectrum averaged across a time-window centered at 90-th frame.}
        \label{fig:spectrum 90th frame}

\end{minipage}
\begin{minipage}[T]{0.40\textwidth}
\centering
    \vspace{+3mm}
    \begin{tabular}{lcc} 
    \toprule
    Time horizon & 6:12 & 13:30 \\ 
    \midrule
    FNO & 1.24 & 1.61 \\
    TFNO \citep{kossaifi2023tfno} & 1.25 & 1.81 \\
    UNet & 1.65 & 4.66 \\
    CNextUNet \citep{ohana2025well} & 1.30 & 2.23 \\ 
    \midrule
    Latent-AR & 0.89 & 1.25 \\
    Latent-FM (ens=1) & 0.88 & 1.24 \\
    Latent-FM (ens=8) & \textbf{0.87} & \textbf{1.24} \\
    \bottomrule
    \end{tabular}
    \vspace{+5mm}
    \captionof{table}{NRMSE for different model's prediction on MHD under different rollout length.    \label{tab: nrmse mhd}}
\end{minipage}  
\end{center}

\begin{figure}[h]
    \centering
    \includegraphics[width=\linewidth]{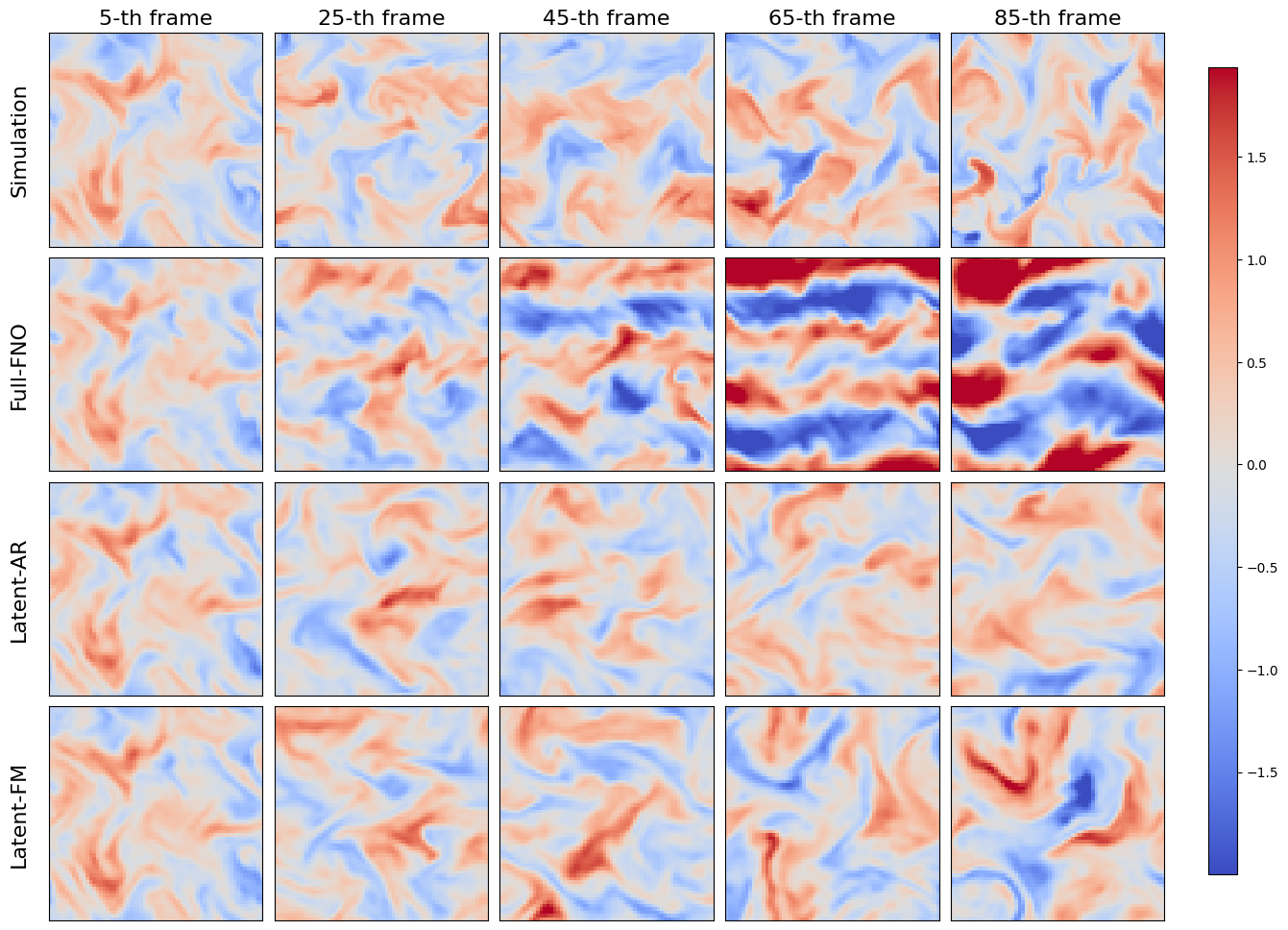}
    \caption[MHD velocity rollout visualization]{Visualization of the velocity field ($x$ component) in 3D MHD predicted by different models.}
    \label{fig:mhd rollout vis}
    \vspace{-4mm}
\end{figure}

For the 3D compressible MHD problem, we compare against the benchmark reported in the Well \citep{ohana2025well}, which comprises FNO, Tensorized-FNO (TFNO) \citep{kossaifi2023tfno}, UNet and a modified version of UNet using ConvNext block \citep{liu2022convnext}. Compared to 2D problems, 3D turbulence exhibit more complex energy cascade due to vortex stretching. The models will need to resolve the spatiotemporal features of all scales to accurately predict the characteristics of the field. As shown in Table \ref{tab: nrmse mhd}, all the models' predictions quickly de-correlate with the reference numerical simulation after a short time period. Similar patterns have been observed in \citet{lienen2024zeroturb}, where it is infeasible for autoregressive neural solver to produce accurate long-term forecast on the chaotic 3D turbulence. Nonetheless, we are still interested at studying the behavior of model in terms of the system states it generated. To this end, we compare the spectrum of prediction generated by FNO that operates on the full-order space and the deterministic/diffusion models that operate on the reduce-order space. A qualitative comparison is shown at Figure \ref{fig:mhd rollout vis}. We can see that all models' predictions deviate from the reference numerical simulation after 5-th frame. For the latent space models, their predictions tend to be more physically coherent in the long run. We further look at the spectrum of different models' prediction (Figure \ref{fig:spectrum 10th frame}, \ref{fig:spectrum 50th frame}, \ref{fig:spectrum 90th frame}). All the models' predictions match reference spectrum in low-frequency regime at the beginning, and FNO is better at the high-frequency regime due to the mesh-reduced nature of the latent space models. As time evolves, the error in thehigher frequency regime of FNO's prediction gradually transport to the low frequency regime, resulting in a distribution shift. For the latent space model, the error accumulation in the spectrum is less pronounced and it can be observed that after a long rollout, the flow matching model generates more coherent spectrum.

\begin{table}[h!]
\centering
\fontsize{9pt}{10pt}\selectfont
\begin{tabular}{lcccccc} 
\toprule
Time horizon & \multicolumn{2}{c}{+1} & \multicolumn{2}{c}{+50} & \multicolumn{2}{c}{+250} \\ 
\cmidrule(lr){1-1}\cmidrule(lr){2-3}\cmidrule(lr){4-5}\cmidrule(lr){6-7}
Variables & $\mathbf{u}$ & $\mathbf{p}$ & $\mathbf{u}$ & $\mathbf{p}$ & $\mathbf{u}$ & $\mathbf{p}$ \\ 
\midrule
MeshGraphNet \citep{pfaff2021mgn} & 0.0810 & 0.4256 & 0.5926 & 2.2492 & 1.0702 & 3.7220 \\
EAGLE Transformer \citep{janny2023eagle} & \textbf{0.0537} & 0.4590 & \textbf{0.3494} & 1.4432 & 0.6826 & 2.4130 \\ 
\midrule
Latent-AR & 0.0766 & 0.0756 & 0.4216 & 0.4157 & 0.6678 & 0.5926 \\
Latent-FM (ens=1) & 0.0724 & \textbf{0.0720} & 0.4030 & 0.4115 & 0.6625 & 0.6077 \\
Latent-FM (ens=8) & 0.0765 & 0.0755 & 0.3925 & \textbf{0.3929} & \textbf{0.6360} & \textbf{0.5795} \\
\midrule
\end{tabular}
\vspace{-2mm}
\caption[NRMSE of EAGLE]{NRMSE for different model's prediction on EAGLE dataset. MeshGraphNet and EAGLE Transformer are trained with backprop-through-time on eight consecutive time steps.}
\label{tab:eagle nrmse}
\vspace{-2mm}
\end{table}%
\begin{figure}[h!]
\centering
\vspace{-2mm}
    \begin{subfigure}{\textwidth}
        \includegraphics[width=\linewidth]{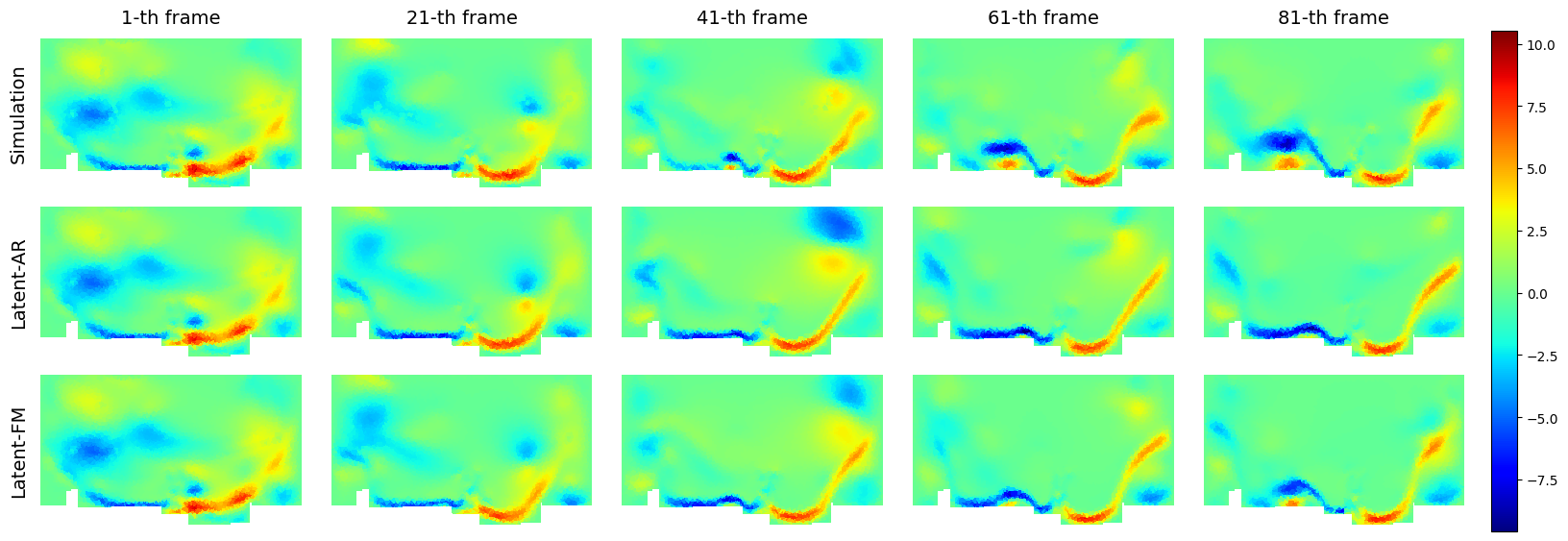}
        \caption{Visualization of different models' prediction on the $x$ component of velocity field.}
    \end{subfigure}
    \begin{subfigure}{\textwidth}
        \vspace{+4mm}
        \includegraphics[width=\linewidth]{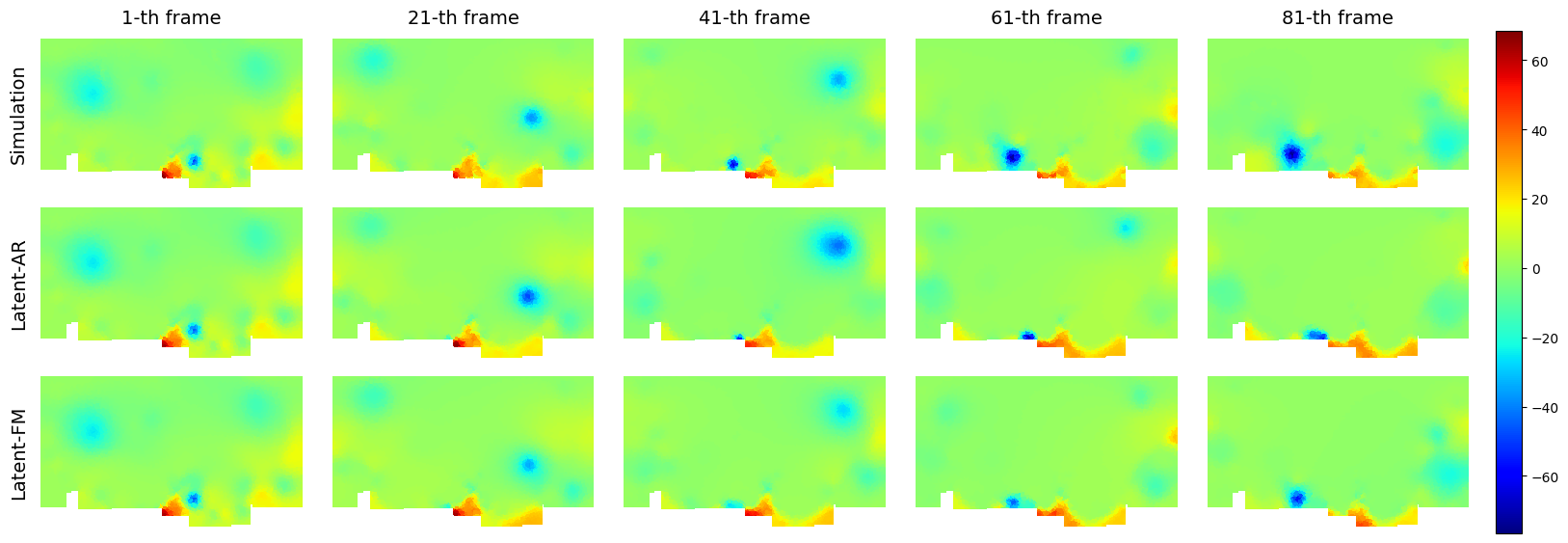}
        \caption{Visualization of different models' prediction on the $x$ component of pressure field.}
    \end{subfigure}
    \caption{Rollout visualization on EAGLE dataset.}
    \vspace{-3mm}
\end{figure}

The third problem is the 2D airflow around the UAV, which comprises a variety of different geometries and features fluid flow coupled with a moving UAV. Previous methods build graph neural networks (GNN) based on the simulation meshes, i.e. a graph with connectivity the same as the discretization mesh. In contrast, our proposed model is built on a unified latent grid. The quantitative comparison between different models are shown in Table \ref{tab:eagle nrmse}. Overall, GNN-based models achieve slightly better accuracy in velocity prediction whereas the proposed latent flow matching model has significantly better accuracy in pressure prediction. We also find that the ensemble prediction notably improves the RMSE of model's prediction in this problem. As shown in the qualitative comparison between deterministic latent models and stochastic ones, we observe that the proposed latent flow matching models can generate predictions that are more physically coherent with the reference numerical simulation.


\section{Conclusion}

In this work, we introduce a neural PDE solver that pairs diffusion-based training/sampling with a dimension-reduction autoencoder, projecting inputs from diverse meshes onto a lower-resolution latent grid. We analyze existing diffusion paths and proposed a modified flow-matching schedule that empirically outperforms other noise schedules. By leveraging multi-step diffusion—reminiscent of classical multi-step methods, our approach mitigates spectral artifacts and reduces compounding errors in chaotic, time-dependent systems. This framework not only enhances temporal stability but also lowers computational cost, offering robust, flexible modeling across various discretization grids. Overall, our generative latent PDE solver presents a scalable and adaptable alternative for designing temporal neural surrogate models.


\printbibliography



%
\end{document}


\appendix
\renewcommand{\thefigure}{S\arabic{figure}}
\renewcommand{\thetable}{S\arabic{table}}

\setcounter{figure}{0}
\setcounter{table}{0}

\rule[0pt]{\columnwidth}{1pt}
\begin{center}
    \huge
    \emph{Supplementary Material} \\
    {\huge {CaFA:} Global Weather Fore\textbf{Ca}sting \\ with \textbf{F}actorized \textbf{A}ttention on Sphere}
\end{center}
\rule[0pt]{\columnwidth}{1.5pt}

\section{Model implementation}
\subsection{Training details}
\label{appendix model training details}

We specify the detailed hyperparamters for the model training in Table \ref{tab:train hyperparameter stage 1} and \ref{tab:train hyperparameter stage 2}. We implement the model and the experiment using PyTorch \citep{paszke2019pytorch}. We use AdamW optimizer \citep{adam15, adamw2018} and the majority of the hyperparameter setting follows ClimaX \citep{climax2023icml}. Regularization techniques including dropout \citep{dropout2014} and droppath \citep{larsson2016fractalnet} are not used in the proposed model (which is used in ClimaX baseline). For testing, we evaluate the final model checkpoint and no early stopping is applied. During second-stage training, gradient checkpointing is used to reduce the memory cost of backpropagating through time. In addition, only the most recent 20 years' (up to the validation years) training data is used for second stage training.

\begin{table}[H]
\centering
\begin{tblr}{
  cells = {c},
  hline{1} = {-}{0.08em},
  hline{2} = {-}{0.05em},
  hline{12} = {-}{0.08em},
}
\textbf{Training Hyperparameter}             & \textbf{Value} \\
Optimizer          & AdamW            \\
Learning rate     & $1e-8 \mapsto 3e-4 \mapsto 1e-7$          \\
Batch size   & 8              \\
Gradient steps           & 160k            \\
Learning rate scheduler  & Cosine Annealing with \\ & Linear Warmup (for 1600 steps)            \\
Weight decay          & 1e-6             \\
Rollout step curriculum & $[1, 2]$          \\
Curriculum milestone & $[0, 50\text{k}] $           \\
EMA & None
\end{tblr}
\caption{\label{tab:train hyperparameter stage 1} Training hyperparameter for stage 1. On $64\times32$ grid we adjust the batch size to $16$ and peak learning rate to $5e-4$. The curriculum milestone determines how many steps to rollout after specific milestone. In the above setting, after 50k gradient steps, the model will rollout for 2 steps.}
\end{table}

\begin{table}[H]
\centering
\begin{tblr}{
  cells = {c},
  hline{1} = {-}{0.08em},
  hline{2} = {-}{0.05em},
  hline{12} = {-}{0.08em},
}
\textbf{Training Hyperparameter}             & \textbf{Value} \\
Optimizer          & AdamW            \\
Learning rate     & $1e-8 \mapsto 3e-7 \mapsto 1e-7$          \\
Batch size   & 8              \\
Gradient steps           & 50k            \\
Learning rate scheduler  & Cosine Annealing with \\ & Linear Warmup (for 500 steps)            \\
Weight decay          & 1e-6             \\
Rollout step curriculum & $[4, 8, 12, 16, 20]$          \\
Curriculum milestone & $[0, 10\text{k}, 20\text{k}, 30\text{k}, 40\text{k}] $           \\
EMA & Deacy=0.999
\end{tblr}
\caption{\label{tab:train hyperparameter stage 2} Training hyperparameter for stage 2. We save the Exponential Moving Average (EMA) of model weights during the second stage.}
\end{table}

The training loss is a latitude-weighted and variable-weighted $L1$ loss:
\begin{equation}
    \label{eq:training loss}
    \text{Loss}(\hat{X}, X) =
     \frac{1}{N_{\text{lat}}N_\text{lon}N_\text{t}}
    \sum_{c=1}^{N_c}\lambda_c
    \sum_{l=1}^{N_\text{level}}\lambda_l
    \sum_{t=1}^{N_{\text{t}}}
    \sum_{i=1}^{N_{\text{lat}}}
    \sum_{j=1}^{N_{\text{lon}}}
    w(i)||\hat{X}_{tijl}^c-X_{tijl}^c||
    ,
\end{equation}
where $w(i)$ is the latitude-based weight,  $\lambda_l$ is the level-based weight, $\lambda_c$ is the variable-based weight,  $\hat{X}$ is model's prediction and $X$ is reference data, $N_t$ denotes total number of rollout timesteps, $N_{\text{lat}}$ denotes number of latitude grid points, $N_{\text{lon}}$ denotes number of longitude grid points. The level-based weight is a simple linear interpolation between $0.05$ and $0.065$: $\lambda_l=\text{Index}(l)/12 \times 0.015 + 0.050$, where the index of pressure level $l$ ranges from 0 to 12 corresponds to 50hPa to 1000 hPa.  We list the specification of the variable-based weight below (see Table 2 in main text for the meaning of the variable abbreviation):

\begin{table}[H]
\centering
\begin{NiceTabular}{lccccccccccc} 
\toprule
\textbf{Variable} & z & t & u & v & q & t2m & MSLP & u10 & v10 & TP6 & TP24 \\
\cmidrule(lr){2-12}
\textbf{Weight} & 1.0 & 1.0 & 0.5 & 0.5 & 0.1 & 1.0 & 0.1 & 0.1 & 0.1 & 0.05 & 0.05 \\
\bottomrule
\end{NiceTabular}
\caption{Variable-based weight. \label{tab:variable weight}}
\end{table}

We didn't perform search over the best combination of variable-based weights and level-based weights. We find uniform level-based weights perform almost similar to current level-based weights in preliminary experiments. The variable weights we've used is similar to the variable weights used in GraphCast\citep{graphcast2023science}. Recent work Stormer \citep{nguyen2023stormer} showcase that using variable-based weight is beneficial for variables that are assigned higher loss weight like 2m temperature, geopotential.

\subsection{Metrics}
\label{appendix metric definition}

The following metrics are defined on a surface variable or a upper-air variable at a specific level.
\paragraph{Root mean squared error (RMSE)}
The RMSE for time step $t$ is defined as:
\begin{equation}
    \label{eq:rmse}
    \text{RMSE}_t(\hat{X}, X) = \sqrt{
    \frac{1}{N_{\text{lat}}N_\text{lon}}
    \sum_{i=1}^{N_{\text{lat}}}
    \sum_{j=1}^{N_{\text{lon}}}
    w(i)(\hat{X}_{tij}-X_{tij})
    }.
\end{equation}

\paragraph{Anomaly correlation coefficient (ACC)} The ACC is defined as the Pearson correlation coefficient of the anomalies with respect to the
climatology:
\begin{align}
    \label{eq:acc}
    &\text{Anomaly:} \quad \hat{X}'_{tij} = \hat{X}_{tij} - C_{\tilde{t}ij}, \quad  X'_{tij} = \hat{X}_{tij} - C_{\tilde{t}ij}, \\
    &\text{ACC}_t(\hat{X}', X') = \frac{
     \sum_{i=1}^{N_{\text{lat}}}
    \sum_{j=1}^{N_{\text{lon}}}w(i)\hat{X}'_{tij}X'_{tij}}
    {\sqrt{ 
    \left(\sum_{i=1}^{N_{\text{lat}}}
    \sum_{j=1}^{N_{\text{lon}}}w(i)\hat{X'}^2_{tij}\right)
     \left(\sum_{i=1}^{N_{\text{lat}}}
    \sum_{j=1}^{N_{\text{lon}}}w(i)X^2_{tij}\right)}},
\end{align}
where we use the climatology statistics $C$ provided in WeatherBench2 \citep{rasp2023weatherbench2} (computed on ERA5 data from $1990$ to $2019$), $\tilde{t}$ denotes the day of year and time of day of the corresponding time index $t$.

\paragraph{Bias} The bias is defined as the point-wise difference between prediction and ground truth:
\begin{equation}
     \label{eq:bias}
    \text{Bias}_t(\hat{X}, X) = \hat{X}_{tij} - X_{tij}.
\end{equation}%
\section{Comparison against standard attention}
\label{appendix attention comparison}
\begin{figure}[H]
    \centering
    \includegraphics[width=0.9\linewidth]{appendix_figures/Diagram_factorized_attention.pdf}
    \caption{Axial factorized attention. The multi-dimensional spatial structure of value is preserved, each axial attention kernel matrix's size is $S_m^2$ for $n$-dimensional grid with total size $ S=S_1\times S_2\times \hdots \times S_n$. }
    \label{fig:diagram factorized attention}
    \vspace{2mm}
    \includegraphics[width=0.9\linewidth]{appendix_figures/Diagram_full_attention.pdf}
    \caption{Standard attention for sequence. The multi-dimensional spatial structure is not considered explicitly during the attention (it can be implicitly encoded via positional encoding). The full attention kernel matrix's size is quadratic to the total grid size $S$.}
    \label{fig:diagram attention}
\end{figure}

Two illustrative diagrams of factorized attention and standard attention are shown above (Figure \ref{fig:diagram factorized attention} and \ref{fig:diagram attention}). Instandard attention, the spatial structure is not preserved during the contraction of value and attention kernel. Compared to axial Transformer \citep{ho2020axial}, the axial attention kernel in CaFA is still dependent on global information while in axial Transformer its context is limited to that specific row/column. 

We compare factorized attention to two different implementations of standard attention in PyTorch's attention function: \verb|torch.nn.functional.scaled_dot_product_attention|
- a standard one without memory optimization and another one with memory optimization \citep{rabe2021memefficient} \footnote {FlashAttention \citep{flashattention2022nips} is currently not available for \texttt{float32} computation on A6000 in PyTorch 2.1.}. The results are shown in Figure \ref{fig:latency}, \ref{fig:flops}, \ref{fig:memory}. We observe that factorized attention has significantly better scaling efficiency in terms of runtime and FLOPs. At the lowest resolution, the runtime of factorized attention is larger than standard attention. This is because in factorized attention layer the input is first processed by two projection layer and then sent to compute the axial attention kernels, with all computation done sequentially  \footnote{For projection and attention kernel's computation, each axis's computation are independent to each other and the memory cost is small, so potentially they can be optimized to run in parallel.}. Compared to memory efficient attention, factorized attention has larger memory footprint at current stage. The memory cost of factorized attention can also be optimized by adopting a similar chunking strategy in \citet{rabe2021memefficient} during the contraction of value and attention kernels.

\begin{figure}[H]
    \centering
    \vspace{-2mm}
    \begin{subfigure}{\textwidth}
    \centering
    \includegraphics[width=0.6\textwidth]{appendix_figures/latency.png}
    \caption{Forward latency}
    \label{fig:latency}
    \end{subfigure}
    \begin{subfigure}{\textwidth}
        \centering
        \includegraphics[width=0.6\textwidth]{appendix_figures/flops.png}
    \caption{Theoretical FLOPs}
    \label{fig:flops}
    \end{subfigure}
    \begin{subfigure}{\textwidth}
        \centering
        \includegraphics[width=0.6\textwidth]{appendix_figures/memory.png}
    \caption{Peak memory usage}
    \label{fig:memory}
    \end{subfigure}
    \vspace{-2mm}
    \caption{\label{fig:computational benchmark}Computational benchmark between factorized attention and different implementation of the standard softmax attention \citep{attention2017nips}. The number of attention head is 16, each attention head's dimension is 128, batch size is set to 1, input and output channel dimension is 512. The benchmark is carried out on A6000 with CUDA 12, \texttt{float32} precision is used. The runtime statistics are profiled using DeepSpeed. The factorized attention layer contains following learnable modules that are not in standard attention layer: distance encoding and projection layer.}
        \vspace{-2mm}
\end{figure}%
\section{Further results}
\label{appendix more results}

In this section we provide more quantitative and qualitative analysis of model's prediction capability.

We analyze how the model perform in terms of $L1$ and $L2$ (RMSE) loss. The normalized difference with respect to IFS HRES is shown in Figure \ref{fig:l1-l2-diff}. We observe that the model performs relatively better in terms of $L1$ norm in the long run, which is possibly because that the model is trained on $L1$ norm and penalizes outlier less than $L2$ norm. We also compare models trained with different time step sizes (lead time 6 hour vs 12 hour). We find that model with larger time step size performs better in terms of forecasting accuracy beyond 7 days but performs worse in shorter range, while the ACC of both model variants fall below $0.6$ on day 10.

Example visualization of the learned positional encoding, distance encoding and attention kernels are shown in Figure \ref{fig:spherical pe}, \ref{fig:attention kernel visualization}, \ref{fig:distance encoding visualization}. Interestingly, the attention kernel at earlier layer exhibit sharper patterns than deeper layer, where attention scores are mostly concentrated around the diagonal.

\begin{figure}[H]
    \centering
    \includegraphics[width=\textwidth]{appendix_figures/l1_l2_norm.pdf}
    \caption{Relative $L1$ and $L2$ error difference with respect to IFS HRES of 4 selected key variables.}
    \label{fig:l1-l2-diff}
\end{figure}%
\begin{figure}[H]
    \centering
    \includegraphics[width=\textwidth]{appendix_figures/6hvs12h.pdf}
    \caption{Normalized RMSE difference and ACC difference comparison between CaFA trained with $6$ hour interval and CaFA trained with $12$ hour interval. Negative RMSE difference and positive ACC difference indicates better performance. CaFA trained with $12$ hour interval uses less gradient steps for the second stage (36k) and fewer rollout steps (up to 16).}
    \label{fig:6hvs12h}
\end{figure}

\begin{figure}[H]
    \centering
    \includegraphics[width=0.85\textwidth]{appendix_figures/spherical_pe.png}
    \caption{Example visualization of learned spherical harmonic based positional encoding. "Layer" denotes their corresponding attention layer number, "c" denotes channel number.}
    \label{fig:spherical pe}
\end{figure}%
\begin{figure}[H]
    \centering
    \includegraphics[width=0.90\textwidth]{appendix_figures/attention_kernel_example.pdf}
    \caption{Example visualization of learned self-attention kernel along different axes. The kernel matrices are selected from random batch and head. For better clarity all the kernel matrices shown are normalized such that all the elements fall into the range of $[-1, 1]$.}
    \label{fig:attention kernel visualization}
\end{figure}%
\begin{figure}[H]
    \centering
    \includegraphics[width=0.90\textwidth]{appendix_figures/distance_encoding_example.pdf}
    \caption{Example visualization of learned distance encoding along different axes. The distance encoding are selected from random batch, head and channel. For better clarity all the kernel matrices shown are normalized such that all the elements fall into the range of $[-1, 1]$.}
    \label{fig:distance encoding visualization}
\end{figure}%
\begin{figure}[H]
    \centering
    \begin{subfigure}{\textwidth}
    \includegraphics[width=0.94\textwidth]{result_figures/v10-vis.pdf}
    \vspace{-1mm}
    \caption{$v10$m example rollout visualization}
    \end{subfigure}  
    \begin{subfigure}{\textwidth}
    \includegraphics[width=0.94\textwidth]{result_figures/u850-vis.pdf}
    \vspace{-1mm}
    \caption{$u850$ example rollout visualization}
    \end{subfigure}
    \vspace{-1mm}
    \begin{subfigure}{\textwidth}
    \includegraphics[width=0.94\textwidth]{result_figures/v850-vis.pdf}
    \vspace{-1mm}
    \caption{$v850$m example rollout visualization}
    \end{subfigure}
\end{figure}%
\begin{figure}[H]\ContinuedFloat 
    \medskip
    \begin{subfigure}{\textwidth}
    \includegraphics[width=0.94\textwidth]{result_figures/t850-vis.pdf}
    \vspace{-1mm}
    \caption{$t850$ example rollout visualization}
    \end{subfigure}
    \begin{subfigure}{\textwidth}
    \includegraphics[width=0.94\textwidth]{result_figures/q700-vis.pdf}
    \vspace{-1mm}
    \caption{$q700$m example rollout visualization}
    \end{subfigure}
    \begin{subfigure}{\textwidth}
    \includegraphics[width=0.94\textwidth]{result_figures/mslp-vis.pdf}
    \vspace{-2mm}
    \caption{MSLP example rollout visualization}
    \end{subfigure}%
    \vspace{-3mm}
    \caption{\label{fig:more qualitative visualization} More example rollout visualization of model's prediction versus reference ERA5 reanalysis data at different lead times. The initialization time is 00:00 UTC on August 11, 2020. }
    \vspace{-10mm}
\end{figure}

\printbibliography



%